\documentclass[letterpaper]{article} % DO NOT CHANGE THIS
\usepackage[]{aaai23}  % DO NOT CHANGE THIS
\usepackage{times}  % DO NOT CHANGE THIS
\usepackage{helvet}  % DO NOT CHANGE THIS
\usepackage{courier}  % DO NOT CHANGE THIS
\usepackage[hyphens]{url}  % DO NOT CHANGE THIS
\usepackage{graphicx} % DO NOT CHANGE THIS
\usepackage{natbib}  % DO NOT CHANGE THIS AND DO NOT ADD ANY OPTIONS TO IT
\usepackage{caption} % DO NOT CHANGE THIS AND DO NOT ADD ANY OPTIONS TO IT
\usepackage{color}
\usepackage{algorithm}
\usepackage{algorithmic}
\usepackage{amsmath,amssymb}
\usepackage{makecell}
\usepackage{threeparttable}
\usepackage{soul}
\usepackage{newfloat}
\usepackage{listings}
\DeclareCaptionStyle{ruled}{labelfont=normalfont,labelsep=colon,strut=off} % DO NOT CHANGE THIS
\floatstyle{ruled}
\newfloat{listing}{tb}{lst}{}
\floatname{listing}{Listing}

\title{TLDW: Extreme Multimodal Summarisation of News Videos}

\author{
    Peggy Tang, \textsuperscript{\rm 1}
    Kun Hu, \textsuperscript{\rm 1}
    Lei Zhang, \textsuperscript{\rm 2}
    Jiebo Luo, \textsuperscript{\rm 3} 
    Zhiyong Wang \textsuperscript{\rm 1}
}

\affiliations {
    \textsuperscript{\rm 1} School of Computer Science, The University of Sydney \\
    \textsuperscript{\rm 2} International Digital Economy Academy (IDEA) \\
    \textsuperscript{\rm 3} Department of Computer Science, University of Rochester \\
    peggy.tang@sydney.edu.au, kuhu6123@uni.sydney.edu.au, leizhang@idea.edu.cn, \\
    jluo@cs.rochester.edu, zhiyong.wang@sydney.edu.au
}

\begin{document}

\maketitle

\begin{abstract}
Multimodal summarisation with multimodal output is drawing increasing attention due to the rapid growth of multimedia data. 
While several methods have been proposed to summarise visual-text contents, their multimodal outputs are not succinct enough at an extreme level to address the information overload issue.
To the end of extreme multimodal summarisation, we introduce a new task, \textit{eXtreme Multimodal Summarisation with Multimodal Output (XMSMO)} for the scenario of TL;DW - \textit{Too Long; Didn't Watch}, akin to TL;DR. XMSMO aims to summarise a video-document pair into a summary with an extremely short length, which consists of one cover frame as the visual summary and one sentence as the textual summary. We propose a novel {\it unsupervised} Hierarchical Optimal Transport Network (HOT-Net) consisting of three components:  hierarchical multimodal encoders,  hierarchical multimodal fusion decoders, and  optimal transport solvers. 
Our method is trained, without using reference summaries, by optimising the visual and textual coverage from the perspectives of the distance between the semantic distributions under optimal transport plans. 
To facilitate the study on this task, we collect a large-scale dataset XMSMO-News by harvesting 4,891 video-document pairs. The experimental results show that our method achieves promising performance in terms of ROUGE and IoU metrics. \footnote{Our dataset and source code will be publicly available in GitHub.}
\end{abstract}

\section{Introduction}

Summarisation aims to condense a given piece of information into a short and succinct summary that best covers its semantics with the least redundancy. This helps users quickly browse and understand long content by focusing on the most important ideas \cite{IMani2001automatic}. Summarisation on a single modality, such as video summarisation \cite{MMa2020videosparse, LYuan2020unsupervisedvideo}, which aims to summarise a video into keyframes, and text summarisation \cite{RMihalcea2004Textrank, ASee2017Get, YLiu2019BERTExt, PLaban2020SummaryLoop}, which aims to summarise a document into a few sentences, has been actively studied for decades.

Video summarisation aims to summarise a video into keyframes \cite{DBLP:journals/tcsvt/LuoPC09, JWang2019stacked, LYuan2020unsupervisedvideo,JWang2020QueryTwiceVS} that provide a compact yet informative representation of a video. The majority of existing methods focus on modelling the temporal dependency and spatio structure among frames \cite{EApostolidis2021vssurvey}. To address information overload issues, extreme video summarisation has been proposed as a sub-task of video summarisation  \cite{HGu2018Thumbnails, JRen2020BestFrame, EApostolidis2021RLThumbnail}, which aims to summarise a video into a cover frame. It involves high source compression and allows users to quickly discern the essence of a video and decide whether it is worth watching or not. 

\begin{figure}[ht]
\centering
\includegraphics[width=.47\textwidth]{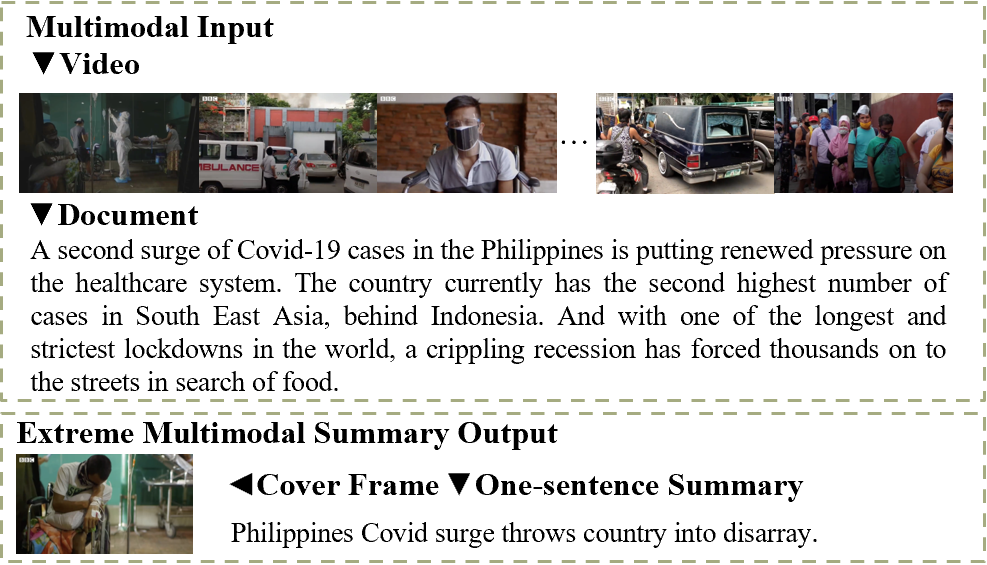}
\caption{Illustration of our newly proposed task XMSMO.}
\label{fig:xmsmo}
\end{figure}

Text summarisation aims to condense a given document into a short and succinct summary that best covers the document's semantics.  
The majority of existing methods are either extractive or abstractive. Extractive methods \cite{SNarayan2018Ranking,XZhang2019HIBERT,YLiu2019BERTExt,zhong2020extractive} select salient sentences from a document to form its summary. Abstractive methods \cite{ASee2017Get, RPaulus2018DeepReinforced, JZhang2020pegasus, PLaban2020SummaryLoop} involve natural language generation to generate a summary for a given document.  
To further condense the text and address information overload issues, extreme text summarisation has been proposed as a sub-task of text summarisation. Extreme text summarisation \cite{SNarayan2018XSum, YLu2020multixscience,ICachola2020tldr, SSotudeh2021tldr9} aims to summarise a document into a one-sentence summary. It helps users quickly browse through the main information of a document.

While single-modal summarisation has been investigated for decades, with the rapid growth of multimedia data, there is an emerging interest  on Multimodal Summarisation with Multimodal Output (MSMO) \cite{JZhu2018msmo,JZhu2020multimodal,MLi2020vmsmo}. MSMO aims to summarise a pair of a video or a set of images and a document into a visual-textual summary, since image and text could complement each other to help users to better obtain a more informative and visual understanding of events.  However, most of the existing MSMO methods are designed for short visual inputs, such as short videos and multiple images, without considering the summary length. Given the increasing pace of producing multimedia data  and the subsequent challenge in keeping up with the explosive growth of such rich content, these existing methods may be sub-optimal to address the imminent issue of information overload of multimedia data. 

In this paper,  we introduce a new task, \textit{eXtreme Multimodal Summarisation with Multimodal Output (XMSMO)}, for the scenario TLDW which stands for \textit{Too Long; Didn't Watch)}. 
As shown in Figure \ref{fig:xmsmo}, XMSMO aims to summarise a pair of a video and its corresponding document into a multimodal summary with an extremely short length. That is, an extreme multimodal summary consists of one cover frame as the visual summary and one sentence as the textual summary. 
To solve this new task, we propose a novel unsupervised Hierarchical Optimal Transport Network (HOT-Net) architecture including three components, the hierarchical multimodal encoders, the hierarchical multimodal (fusion-based) decoders and the optimal transport solvers. 

Specifically, the hierarchical visual encoder formulates the representations of a video from three levels including frame-level, scene-level and video-level; the hierarchical textual encoder formulates the representations of a document from three-levels as well: word-level, sentence-level and document-level. 
Then, the hierarchical decoder formulates the cross-modal representations in a local-global manner and evaluates candidate cover frames and candidate words, which are used to form a visual summary and a compressive textual summary, respectively. Note that a compressive textual summary offers a balance between the conciseness issue of extractive summarisation and the factual hallucination issue of abstractive summarisation.
%The hierarchical multimodal pooling mechanism attends and fuses the fine and coarse textual and visual context representations to form hierarchical multimodal attention. Multimodal pooling layer are based on graph attentions, since this formulation helps easily extend to additional modalities in the future, such as an audio modality. Each feature of a modality is treated as a vertex feature of a graph. The relationships between modalities are formulated by self-attentions to attend across modalities. The hierarchical visual decoder selects an optimal frame as an extreme visual summary, and the hierarchical textual decoder produces an extreme textual summary. Our extreme textual summary is compressive-based, since it offers a balance between the conciseness issue of extractive summarisation and the factual hallucination issue of abstractive summarisation.
Finally, our optimal transport-based unsupervised training strategy is devised to mimic human judgment on the quality of an extreme multimodal summary in terms of the visual and textual coverage. The coverage is measured by a Wasserstein distance with an optimal transport plan measuring the distance between the semantic distributions of the summary and the original content. In addition, textual fluency and cross-modal similarity are further considered, which can be important to obtain a high quality multimodal summary. %\cite{JZhu2020multimodal}. 

%textual fluency, and cross-modal similarity. We devise loss functions based on these four quality criteria to train our model.  Visual coverage of the cover frame and textual coverage of the one-sentence summary measure the essential quality of a good summary since the summarisation task aims to maximise the semantic coverage of the video-document pair in the output summary, given a length constraint. 
%The cross-modal similarity between the cover frame and the one-sentence summary is the crucial quality of a good piece of multimodal summary \cite{JZhu2020multimodal}, and could be a challenge for TLDW. 

Additionally, to facilitate the study on this new task XMSMO and evaluate our proposed HOT-Net, we built the first dataset of such kind, namely XMSMO-News, by harvesting 4,891 video-document pairs as input and cover frame-title pairs as multimodal summary output from the British Broadcasting Corporation (BBC) News Youtube channel from year 2013 to 2021.

In summary, the key contributions of this paper are:
\begin{itemize}
    \item We introduce a new task, eXtreme Multimodal Summarisation with Multiple Output (XMSMO) as TLDW, which stands for \textit{Too Long; Didn't Watch}. It aims to summarise a video-document pair into an extreme multimodal summary (i.e., one cover frame as the visual summary and one sentence as the textual summary). %, while existing extreme summarisation studies are unimodal based.
    \item We propose a novel unsupervised Hierarchical Optimal Transport Network (HOT-Net). The hierarchical encoding and decoding are conducted across both the visual and textual modalities, and optimal transport solvers are introduced to guide the summaries to maximise their semantic coverage. 
%    \item We design a new unsupervised training strategy that mimics the human judgment of multimodal summary quality by minimising the quartet loss of visual coverage, textual coverage, textual fluency, and cross-modal similarity. 
    \item We construct a new large-scale dataset XMSMO-News for the research community to facilitate research in this new direction. Experimental results on this dataset demonstrate that our method outperforms other baselines in terms of ROUGE and IoU metrics.
    
\end{itemize}

\section{Related Work}

%Most of the existing works on extreme summarisation focus on single modality, namely extreme text summarisation and extreme video summarisation. Extreme text summarisation summarises a document into a one-sentence summary. Extreme video summarisation summarises a video by as cover frame. 

In this section, we first review existing deep learning-based extreme unimodal summarisation methods in two categories, video-based and text-based, since they are closely related to our study. We also review existing multimodal summarisation with multimodal output methods which share similar input and output modalities with our study.

\begin{figure*}[h!]
\centering
\includegraphics[width=.99\textwidth]{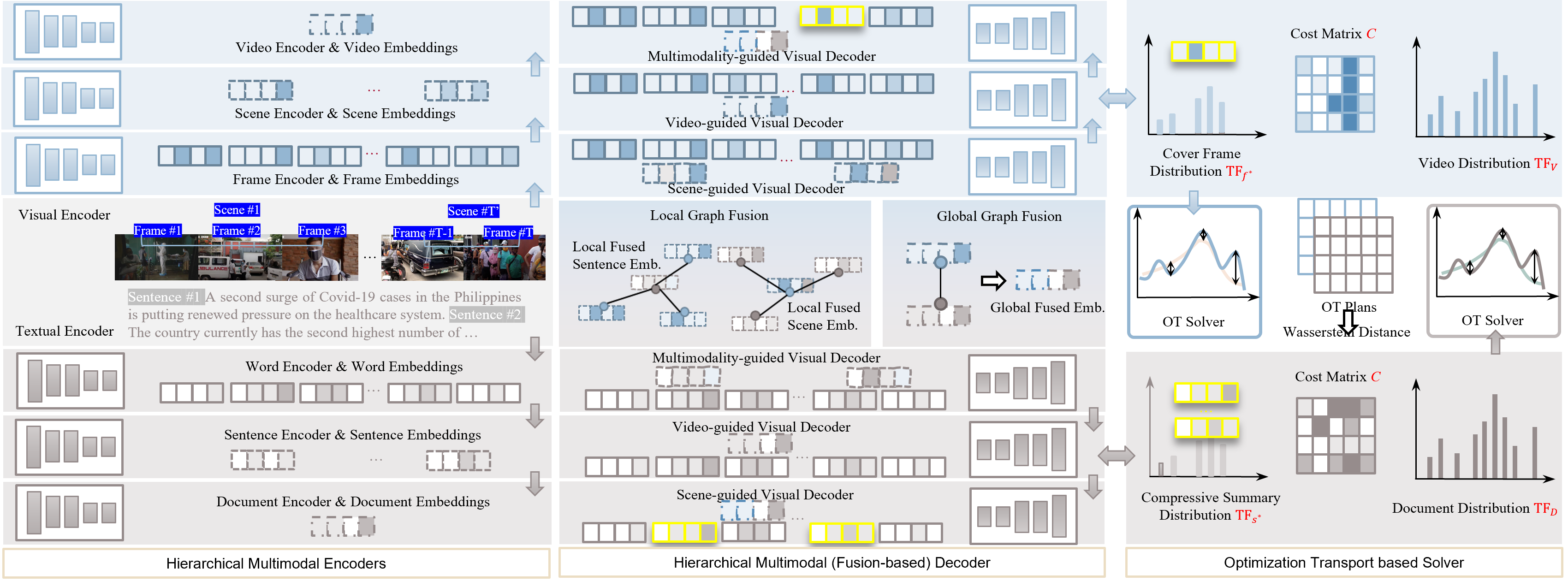}
%a draft version
\caption{Illustration of the unsupervised Hierarchical Optimal Transport Network (HOT-Net) for XMSMO. HOT-Net consists of three components:  hierarchical multimodal encoders,  hierarchical multimodal fusion decoders, and optimal transport solvers.
%; it is trained without reference summaries by optimising the visual and textual coverage 
%from the perspectives of the distance between the semantic distributions 
%under optimal transport plans. 
%The hierarchical encoding and decoding mechanisms are guided by the scene-level and video-level features of the video and the sentence-level and document-level features of the document, and the hierarchical multimodal pooling mechanism attends and fuses the textual and visual context representation to form a multimodal representation.  The model is trained without using reference summaries by minimising the quartet loss of visual coverage, textual coverage, textual fluency, and cross-modal similarity.
}
\label{fig:architecturechart}
\end{figure*}

\subsection{Extreme Video Summarisation} 

Extreme video summarisation methods can be conceptualized as a frame ranking task, which scores the frames in a video as the output.  A deep learning method based on a CNN-based autoencoder architecture was first proposed~\cite{HGu2018Thumbnails}, which was trained by a reconstruction loss considering the representativeness and aesthetic quality of the selected frames. 
The scoring was improved by~\citet{JRen2020BestFrame} by considering the quality of faces, and it utilised a Siamese architecture, which was optimized by a piece-wise ranking loss using pairs of frames. \citet{EApostolidis2021RLThumbnail} proposed a generative adversarial network which introduced a reinforcement learning scheme by rewarding the representativeness and aesthetic quality. Note that most of these methods encode a video as a sequence of frames directly, whilst the semantic hierarchical structure of a video has not been adequately explored.

\subsection{Extreme Text Summarisation} 

%Extreme text summarisation is generally formulated as a sequence-to-sequence generation task, with the source document as the input sequence and the summary as the output sequence.

%Most of the existing methods follow the supervised encoder-decoder framework \cite{SNarayan2018XSum,ICachola2020tldr}. Some methods \cite{SNarayan2018XSum} were proposed to incorporate the topic model as an additional input to the encoder to capture document-level semantic information and guide the summary to have the same theme as the document. Some methods \cite{ICachola2020tldr} were proposed to apply multitask learning and incorporate title generation as a scaffold task to improve the learning of salient information identification. These models naturally learn to integrate knowledge from the training data while generating an abstractive summary. Prior studies showed that these generative models are highly prone to external hallucination, thus may generate contents that are unfaithful to the original document \cite{JMaynez2020Faithfulness}. 

The extreme text summarisation task was first explored by \citet{SNarayan2018XSum} who formulated a sequence-to-sequence learning problem, where the input was a source document and the output was an extreme summary. A supervised encoder-decoder framework was studied and a topic model was incorporated as an additional input to involve the document-level semantic information and guide the summary to be consistent with the document theme.
\citet{ICachola2020tldr} introduced multitask learning and incorporated the title generation as a scaffold task to improve the learning ability regarding the salient information. These methods relied on integrating the knowledge from pre-trained embedding models to generate abstractive summaries. As a result, these generative models are highly prone to external hallucination and it is possible to generate contents unfaithful to the original document, which was shown by  \citet{JMaynez2020Faithfulness}.

%Most of the methods \cite{HGu2018Thumbnails, JRen2020BestFrame, EApostolidis2021RLThumbnail} focus on scoring the representativeness and aesthetic quality of frames, some consider the quality of faces as well \cite{JRen2020BestFrame}. Most of the methods are CNN-based \cite{HGu2018Thumbnails, JRen2020BestFrame, EApostolidis2021RLThumbnail}. In terms of architecture, some methods follow an autoencoder architecture trained by reconstruction loss  \cite{HGu2018Thumbnails}, some follow a generative adversarial network trained by reinforcement learning to reward for the representativeness and aesthetic quality of frames \cite{EApostolidis2021RLThumbnail},  and follow a Siamese model trained by a piece-wise ranking loss using pairs of frames \cite{JRen2020BestFrame}. Most of these methods \cite{HGu2018Thumbnails, JRen2020BestFrame, EApostolidis2021RLThumbnail} encode the video as a sequence of frames without considering the hierarchical structure.

\subsection{Multimodal Summarisation with Multimodal Output}

Multimodal summarisation with multimodal output task was first studied by \citet{JZhu2018msmo}, which took a document and an image set as the input. A supervised attention based encoder-decoder framework was devised. For encoding, a textual encoder and a visual encoder formulate the document and visual representations, respectively. For decoding, a textual decoder generates a textual summary, and a visual decoder selects the most representative image as a visual summary. Additionally, a multimodal attention layer was incorporated to fuse the textual and visual context information. To 
alleviate the modality-bias issue, a multitask learning was applied to jointly consider the two MSMO subtasks: summary generation and text-image relation recognition \cite{JZhu2020multimodal}. A hierarchical intra- and inter-modality correlation between the image and text inputs was studied to enhance the multimodal context representation \cite{LZhang2021hierarchical}. \citet{MLi2020vmsmo} extended visual inputs to short videos, and introduced self-attentions to improve the multimodal context representation. Nonetheless, most of these methods encode the video and document inputs directly without  considering their semantic hierarchical structure. Moreover, these existing methods have been mainly studied in a supervised manner. To the best of our knowledge, our work is the first unsupervised method for MSMO.

%These methods usually follow an attentional encoder-decoder architecture \cite{JZhu2018msmo, JZhu2020multimodal, MLi2020vmsmo, LZhang2021hierarchical}. For the visual input, some of the methods \cite{JZhu2018msmo,JZhu2020multimodal, LZhang2021hierarchical} are designed for multiple images, some are for short video \cite{MLi2020vmsmo}. To encoding, a textual encoder and a visual encoder formulate the document and video/multiple image representation, respectively. For decoding, a textual decoder generates a textual summary, and a visual decoder selects the most representative frame/image as a visual summary. To fuse the textual and visual context information, some methods \cite{JZhu2018msmo, JZhu2020multimodal} proposed to incorporate a multimodal attention layer. Some methods enhanced this context representation by applying a self-attention mechanism \cite{MLi2020vmsmo} and a hierarchical intra- and inter-modality correlation between the multiple-image and document inputs \cite{LZhang2021hierarchical}. However, most of these methods encode the video and document inputs directly without considering the hierarchical structure \cite{MLi2020vmsmo}. Moreover, the existing methods have been mainly studied in a supervised manner \cite{JZhu2018msmo,JZhu2020multimodal, MLi2020vmsmo, LZhang2021hierarchical}. To the best of our knowledge, our work is the first unsupervised method for multimodal summarization with multimodal output.

\section{Methodology}

As shown in Figure \ref{fig:architecturechart}, our proposed eXtreme Multimodal Summarisation method, namely unsupervised Hierarchical Optimal Transport Network (HOT-N), consists of three components, the hierarchical multimodal encoders, the hierarchical multimodal (fusion-based) decoders and the optimal transport solvers. 
Specifically, the hierarchical visual encoder formulates frame-level, scene-level and video-level representations of a video \(\mathbf{V}\). The hierarchical textual encoder formulates word-level, sentence-level and document-level representations of a document \(\mathbf{D}\). Then, the hierarchical visual decoder selects an optimal frame \(\mathbf{f^{*}}\) as an extreme visual summary, and the hierarchical textual decoder produces an extreme textual summary \(\mathbf{s}^{*}\) based on the cross-modal guidance. Finally, the optimal transport solvers conduct unsupervised learning to optimise the encoders and the decoders in pursuit of the best semantic coverage of the obtained summaries.

\subsection{Hierarchical Multimodal Encoders}

\subsubsection{Visual Encoder}

Given an input video $\mathbf{V}$, it can be represented as a sequence of $T$ frames $\{\mathbf{x}_{i}^{\text{frame}}|i=1,...T\}$. By grouping the consecutive frames with similar semantics, the video can be segmented into a sequence of $T'$ scenes $\{\mathbf{x}_{j}^{\text{scene}}|j=1,...,T'\}$, where $\mathbf{x}_{j}^{\text{scene}}$ consists of the video frames from the $i_{j_0}$-th to the $i_{j_1}$-th frame, where $j_0$ indicates the start index of the frame and $j_1$ indicates the end index of the frame for the $j$-th scene in the video. The hierarchical visual encoder learns the scene-level and video-level representations based on $\mathbf{x}_{i}^{\text{frame}}$ and $\mathbf{x}_{j}^{\text{scene}}$, respectively. 

To characterize a video frame $\mathbf{x}_{i}^{\text{frame}}$, a pre-trained neural network can be introduced. The CLIP model \cite{ARadford2021CLIP} is adopted in this study since it is the state-of-the-art multi-modal embedding model. For the sake of convenience, we use the the symbol $\mathbf{x}_{i}^{\text{frame}}$ to represent this pre-trained feature of the $i$-th frame. To further model the scene-level features, a pooling method is introduced, which is denoted as a function $g^{\text{scene}}$. In detail, for the $j$-th scene, its representation $\mathbf{x}_{j}^{\text{scene}}$ can be obtained by observing its associated frame-level features $\mathbf{x}_{i}^{\text{frame}}$, $i=i_{j_0},..., i_{j_1}$ as:
\begin{equation}
\mathbf{x}_{j}^{\text{scene}} = g^{\text{scene}}(\{\mathbf{x}_{i_{j_0}}^{\text{frame}},...,\mathbf{x}_{i_{j_1}}^{\text{frame}}\}). 
\end{equation}
Particularly, a generalized pooling operator (GPO) \cite{JChen2021GPO} is adopted as the pooling method in this study, since it is shown to be an effective and efficient pooling strategy for different features. 
%into features $\{\mathbf{xf}_{i}|i=10,...,m\}$  by \(\mathbf{xf}_{i} = CLIP(\mathbf{f}_{i})\), where $CLIP()$ denotes the pre-trained CLIP model \cite{ARadford2021CLIP}. We choose CLIP because it is the state-of-the-art pre-trained multi-modal embedding model.  
%To model the hierarchical structure of scene and video, the sequence of frame features are fed into the scene Generalized Pooling Operator (GPO) \cite{JChen2021GPO} to produce an output representation of the scenes \(\{\mathbf{x}_{1}^{\text{scene}},...,\mathbf{x}_{T'}^{\text{scene}}\}\):
%\begin{equation}
%\{\mathbf{x}_{1}^{\text{scene}},...,\mathbf{x}_{T'}^{\text{scene}}\} = \textup{GPO}(\mathbf{xf}_{i})), i \in \left[ 1, M \right]\; ,
%\end{equation}
With the scene-level features, a pooled global (i.e., video-level) representation can be derived as:
\begin{equation}
\mathbf{x}^{\text{video}}  = g^{\text{video}}(\{\mathbf{x}_{1}^{\text{scene}},...,\mathbf{x}_{T'}^{\text{scene}}\}),
\end{equation}
where $g^{\text{video}}$ is a video-level pooling function based on a GPO operator. 
%and the global video transformer encoder to produce an output representation of the video \(\mathbf{x}^{\text{video}}\):
%\begin{equation}
%\mathbf{x}^{\text{video}} = \textup{GPO}(\mathbf{xf}_{i})), i \in \left[ 1, m \right]\; .
%\end{equation}

%\subsection{Hierarchical Textual Encoder}
\subsubsection{Textual Encoder}

A document $\mathbf{D}$ can be viewed as a sequence consisting of $U$ words as $\{\mathbf{x}_{m}^{\text{word}}|m=1,...,U\}$ or a sequence of $U'$ sentences $\{\mathbf{x}_{n}^{\text{sentence}}|n=1,...,U'\}$. The $n$-th sentence consists of consecutive words in $\mathbf{D}$ from the $m_{n_0}$-th to the $m_{n,1}$-th word. Similar to the visual encoder, a hierarchical textual encoder is introduced to learn the sentence-level and the document-level representation.

A pre-trained CLIP model is introduced to formulate the word-level features, which is denoted as $\mathbf{x}_{m}^{\text{word}}$ for the $m$-th word. Next, a pooling mechanism $g^{\text{sentence}}$ is adopted to formulate the sentence-level features. In detail, the $n$-th sentence-level features can be computed as:
\begin{equation}
\mathbf{x}_{n}^{\text{sentence}} = g^{\text{sentence}}(\{\mathbf{x}_{m_{n_0}}^{\text{word}},...,\mathbf{x}_{m_{n,1}}^{\text{word}}\}). 
\end{equation}
Finally, the global representation of the document $\mathbf{D}$ can be derived based on the sentence-level features:
\begin{equation}
\mathbf{x}^{\text{document}} = g^{\text{document}}(\{\mathbf{x}_{1}^{\text{sentence}},...,\mathbf{x}_{U'}^{\text{sentence}}\}),
\end{equation}
where $g^{\text{document}}$ is a document-level pooling function based on GPO. 
%text GPO encoders to produce an output representation of the document \(\mathbf{x}^{\text{video}}\):
%\begin{equation}
%\mathbf{x}^{\text{document}} = \textup{GPO}(\{\mathbf{x}_{1}^{\text{sentence}},...,\mathbf{x}_{U'}^{\text{sentence}}\}_{a})), a \in \left[ 1, n \right]\; .
%\end{equation}

%We encode its words into features $\{\mathbf{xw}_{b}|b=1,...,n\}$  by \(\mathbf{xw}_{b} = CLIP(\mathbf{sw}_{b})\), where $CLIP()$ denotes the pre-trained CLIP model. Then the sequence of word embeddings are fed into the sentence transformer encoder to produce an output representation of the sentence \(\{\mathbf{x}_{1}^{\text{sentence}},...,\mathbf{x}_{U'}^{\text{sentence}}\}\) :
%\begin{equation}
%\{\mathbf{x}_{1}^{\text{sentence}},...,\mathbf{x}_{U'}^{\text{sentence}}\} = \textup{TransformerEncoder}(\mathbf{xw}_{b}), b \in \left[ 1, N \right]\; ,
%\end{equation}
%and the global textual transformer encoders to produce an output representation of the document \(\mathbf{x}^{\text{video}}\):
%\begin{equation}
%\mathbf{x}^{\text{document}} = \textup{TransformerEncoder}(\mathbf{xw}_{a}), a \in \left[ 1, n \right]\; .
%\end{equation}

\subsection{Hierarchical Multimodal Fusion}

To attend and fuse the representations from the visual and textual modalities, we adopt a graph-based attention mechanism~\cite{PVelivckovic2018GAT}. This formulation helps easily extend the attention layer to future additional modalities, such as an audio modality. Each modality feature can be treated as a vertex feature of a graph. The relationships between modalities are formulated by graph convolution to attend over the other modality, which then updates the representations of each modality. Particularly, a hierarchical local, which focuses between scene and sentence levels, and global, which focuses between video and document levels, observations are introduced by a graph fusion strategy.

For local multimodal fusion, the representations of the scenes \(\mathbf{x}^{\text{scene}}=\{\mathbf{x}_{1}^{\text{scene}},...,\mathbf{x}_{T'}^{\text{scene}}\}\) and sentences \(\mathbf{x}^{\text{sentence}}=\{\mathbf{x}_{1}^{\text{sentence}},...,\mathbf{x}_{U'}^{\text{sentence}}\}\) are fed into graph fusion modules $f^\text{scene}_{\text{local}}$ and $f^\text{sentence}_{\text{local}}$. The resulted representation, which can be viewed as an information exchange between modalities, are fed into an average pooling operator $g^{\text{avg}}$ to obtain the local multimodal context representations $\dot{\mathbf{x}}_{j}^{\text{scene}}$ and $\dot{\mathbf{x}}_{n}^{\text{sentence}}$:
\begin{equation}
\begin{aligned}
\dot{\mathbf{x}}_{j}^{\text{scene}} = & g^{\text{avg}}([f^{\text{scene}}_{\text{local}}(\mathbf{x}_{j}^{\text{scene}}; \mathbf{x}_{1}^{\text{sentence}}  )  ,..., \\ & f^{\text{scene}}_{\text{local}}(\mathbf{x}_{j}^{\text{scene}};\mathbf{x}_{U'}^{\text{sentence}} ) ]),
\end{aligned}
\end{equation}
\begin{equation}
\begin{aligned}
\dot{\mathbf{x}}_{n}^{\text{sentence}} = & g^{\text{avg}}([f^{\text{sentence}}_{\text{local}}(\mathbf{x}_{n}^{\text{sentence}}; \mathbf{x}_{1}^{\text{scene}}  )  ,..., \\ & f^{\text{sentence}}_{\text{local}}(\mathbf{x}_{n}^{\text{sentence}};\mathbf{x}_{T'}^{\text{scene}} ) ]).
\end{aligned}
\end{equation}
For global multimodal fusion, the global representations of the document \(\mathbf{x}^{\text{document}}\) and video \(\mathbf{x}^{\text{video}}\) are fed into a graph fusion module $f_{\text{global}}$:
\begin{equation}
\begin{aligned}
%\left[\mathbf{x'}^{\text{video}} ; \mathbf{x'}^{\text{document}}  \right] &=
\dot{\mathbf{x}} = g^{\text{avg}}(f_{\text{global}}(\left[\mathbf{x}^{\text{video}} ; \mathbf{x}^{\text{document}}  \right])).
\end{aligned}
\end{equation}

\subsection{Hierarchical Multimodal Decoders}

\subsubsection{Visual Decoder}
Our visual decoder consists of three stages: 1) scene-guided frame decoding, 2) video-guided frame decoding, and 3) cross-modality-guided frame decoding. It aims to evaluate the probability of a particular frame being a cover frame. 

To produce a scene-aware decoding outcome of evaluating each frame, a scene-guided visual decoder $h^{\text{scene}}$ derives a latent decoding
$\mathbf{y}_{j}^{\text{scene}}$ for frames from $i_{j_0}$ to $i_{j_1}$, $j=1,...,T'$, as follows: 
\begin{equation}
\begin{aligned}
\mathbf{y}_{j}^{\text{scene}} & =  \{\mathbf{y}_{i_{j_0}}^{\text{scene-frame}},...,\mathbf{y}_{i_{j_1}}^{\text{scene-frame}}\} \\ & = h^{\text{scene}}(\{\mathbf{x}_{i_{j_0}}^{\text{frame}},...,\mathbf{x}_{i_{j_1}}^{\text{frame}}\}|\dot{\mathbf{x}}_{j}^{\text{scene}}),
\end{aligned}
\end{equation}
where $h^{\text{scene}}$ is a bi-directional GRU \cite{DBahdanau2015gru} and $\dot{\mathbf{x}}_{j}^{\text{scene}}$ is a multimodal scene guidance, which can be viewed as a prior knowledge.
%\begin{equation}
%\begin{aligned}
%\mathbf{y}_{j}^{\text{scene}} = \textup{Bi-GRU}(input,hidden)\;.
%\end{aligned}
%\end{equation}
Next, 
to produce a video-guided frame decoding outcome, we have:
\begin{equation}
\begin{aligned}
\mathbf{y}^\text{{video}} & = \{\mathbf{y}_{1}^{\text{video-frame}},...,\mathbf{y}_{T}^{\text{video-frame}}\} \\ & = h^{\text{video}}(\{\mathbf{x}_{i_{j_0}}^{\text{frame}},...,\mathbf{x}_{i_{j_1}}^{\text{frame}}\}|\mathbf{x}^{\text{video}}),
\end{aligned}
\end{equation}
where $h^{\text{video}}$ is a bi-directional GRU and $\mathbf{x}^{\text{video}}$ is a unimodal video guidance as a prior knowledge. 
%we utilize the video Bi-GRU decoder to obtain \(\mathbf{y}^\text{{video}}= \left \{ \mathbf{y}^\text{{video}}_{1},..., \mathbf{y}^\text{{video}}_{T} \right \}\) by defining \(input = \{\mathbf{y}_{1}^{\text{scene}},...,\mathbf{y}_{T}^{\text{scene}}\}\), \(hidden=\mathbf{x}^{\text{video}}\),
%\begin{equation}
%\begin{aligned}
%\mathbf{y}^\text{{video}}= \textup{Bi-GRU}(input,hidden)\;.
%\end{aligned}
%\end{equation}
Finally, to produce a global multimodal context-aware decoding, we adopt a Bi-GRU decoder $\dot{h}^\text{video}$ with the guidance of the cross-modal embedding $\dot{\mathbf{x}}$:
\begin{equation}
\begin{aligned}
\dot{\mathbf{y}}^\text{{video}} & = \{\dot{\mathbf{y}}_{1}^{\text{video-frame}},...,\dot{\mathbf{y}}_{T}^{\text{video-frame}}\} \\ & = \dot{h}^\text{video}(\mathbf{y}_{1}^{\text{video-frame}},...,\mathbf{y}_{T}^{\text{video-frame}}|\dot{\mathbf{x}}).
\end{aligned}
\end{equation}
To this end, the optimal frame \(\mathbf{f^{*}}\) is obtained with a frame-wise linear layer activated with a softmax function:
\begin{equation}
\begin{aligned}
\mathbf{f^{*}}  = \textup{argmax}_t(\textup{Linear}(\dot{\mathbf{y}}^\text{{video}})).
\end{aligned}
\end{equation}

\subsubsection{Textual Decoder}
Similar to the visual decoder, the textual decoder also consists of three stages: 1) sentenced-guided word decoding, 2) document-guided word decoding, and 3) cross-modality-guided word decoding. It aims to evaluate the probability of a word being selected in a compressive summary. 

To produce a sentence-aware decoding outcome, a sentence decoder $h^{\text{sentence}}$ derives a latent decoding
$\mathbf{y}_{n}^{\text{sentence}}$ for words from $m_{n_0}$ to $m_{n,1}$, $n=1,...,U'$, where $n_0$ indicates the start index of the word and $n_1$ indicates the end index of the word for the $n$-th sentence in the document, as follows: 

\begin{equation}
\begin{aligned}
\mathbf{y}_{n}^{\text{sentence}} & =  \{\mathbf{y}_{m_{n_0}}^{\text{sentence-word}},...,\mathbf{y}_{m_{n_1}}^{\text{sentence-word}}\} \\ & = h^{\text{sentence}}(\{\mathbf{x}_{m_{n_0}}^{\text{word}},...,\mathbf{x}_{m_{n,1}}^{\text{word}}\}|\dot{\mathbf{x}}_{n}^{\text{sentence}}),
\end{aligned}
\end{equation}
where $h^{\text{sentence}}$ is a bi-directional GRU and $\dot{\mathbf{x}}_{n}^{\text{sentence}}$ is used as a prior knowledge for the multimodal sentence guidance.
Then, to produce a document-level textual decoding, we have:
\begin{equation}
\begin{aligned}
\mathbf{y}^\text{{document}} & = \{\mathbf{y}_{1}^{\text{document-word}},...,\mathbf{y}_{U}^{\text{document-word}}\} \\ & = h^{\text{document}}(\{\mathbf{x}_{m_{n_0}}^{\text{word}},...,\mathbf{x}_{m_{n,1}}^{\text{word}}\}|\mathbf{x}^{\text{document}}),
\end{aligned}
\end{equation}
where $h^{\text{document}}$ is a bi-directional GRU and $\mathbf{x}_{n}^{\text{document}}$ is a unimodal document guidance. 
Finally, to produce a global cross-modal context-aware decoding for each word, a Bi-GRU decoder $\dot{h}^\text{document}$ is adopted with the guidance of the global multimodal embedding $\dot{\mathbf{x}}$:
\begin{equation}
\begin{aligned}
\dot{\mathbf{y}}^\text{{document}} & = \{\dot{\mathbf{y}}_{1}^{\text{document-word}},...,\dot{\mathbf{y}}_{U}^{\text{document-word}}\} \\ & = \dot{h}^\text{document}(\mathbf{y}_{1}^{\text{document-word}},...,\mathbf{y}_{U}^{\text{document-word}}|\dot{\mathbf{x}}).
\end{aligned}
\end{equation}
As a result, the optimal compressive summary \(\mathbf{s}^{*}\) with length \(k\) is obtained by:
\begin{equation}
\begin{aligned}
\mathbf{s}^{*}  = \textup{topk}(\textup{Linear}(\dot{\mathbf{y}}^\text{{document}})).
\end{aligned}
\end{equation}
Note that the selected $k$ words are ranked in line with their scores obtained from the linear layer with a softmax activation. Thus, the sentence $\mathbf{s}^*$ can be constructed with these words and their orders. 

\subsection{Optimal Transport-Guided Semantic Coverage}

Our method is trained without reference summaries by mimicking the human judgment on the quality of a multimodal summary, which minimises a quartet loss of visual coverage, textual coverage, textual fluency, and cross-modal similarity.

\subsubsection{Document Coverage}

Intuitively, a high-quality summary is supposed to be close to the original document regarding their semantic distributions. We measure the Wasserstein distance \cite{MKusner2015WMD} $L_{\text{document}}$ between the document $\mathbf{D}$ and the selected sentence $\mathbf{s}^{*}$. It is the minimal cost required to transport the semantics from $\mathbf{s}^{*}$ to $\mathbf{D}$, measuring the semantic coverage of $\mathbf{s}^{*}$ on $\mathbf{D}$. %$L_{textual\; coverage}$ by computing the Wasserstein distance between the corresponding semantic distributions of the document $\mathbf{D}$ and the summary $\mathbf{S}$. 

Given a dictionary, the number of the $\alpha$-th token (i.e, a word in a dictionary) occurred in $\mathbf{D}$ can be counted as $P_{\mathbf{D}}(\alpha)$. As a result, the semantic distribution $\text{TF}_{\mathbf{D}}$ of the document $\mathbf{D}$ can be defined with the normalized term frequency of each token. In detail, for the $\alpha$-th element of $\text{TF}_{\mathbf{D}}$, we have: 
\begin{equation}
\label{eqt:TFD}
\text{TF}_{\mathbf{D}}(\alpha)=\frac{P_{\mathbf{D}}(\alpha)}{\sum_{\alpha'} P_{\mathbf{D}}(\alpha')}.
\end{equation} 
The semantic distribution $\text{TF}_{\mathbf{s}^{*}}$ of the selected sentence $\mathbf{s}^{*}$ can be derived in a similar manner. The normalized term frequency of the $\alpha$-th token in $\mathbf{s}^{*}$ is: 
\begin{equation}
\label{eqt:TFD}
\text{TF}_{\mathbf{s}^{*}}(\alpha)=\frac{P_{\mathbf{s}^{*}}(\alpha)}{\sum_{\alpha'} P_{\mathbf{s}^{*}}(\alpha')}.
\end{equation}
Note that $\text{TF}_{\mathbf{D}}$ and $\text{TF}_{\mathbf{s}^{*}}$ have an equal total token quantities of \(1\) and can be completely transported from one to the other mathematically. 

A transportation cost matrix $\mathbf{C} = (c_{\alpha\alpha'})$ is introduced to measure the semantic similarity between the tokens. Given a pre-trained tokeniser and token embedding model, define $\mathbf{u}_\alpha$ to represent the feature embedding of the $\alpha$-th token. The transport cost $c_{\alpha\alpha'}$ from the $\alpha$-th token to the $\alpha'$-th one is computed based on the cosine similarity:
%can be written as:
\begin{equation} \label{eqt:costfunction}
c_{\alpha\alpha'} = 1- \frac{<\mathbf{u}_{\alpha}, \mathbf{u}_{\alpha'}>}{\left \| \mathbf{u}_{\alpha} \right \| _{2}\left \|  \mathbf{u}_{t\alpha'} \right \|_{2} }.
\end{equation}
Note that the method to obtain token representations $\mathbf{u}_{\alpha}$ follows the same method that we formulate for word representations $\mathbf{x}_{(\cdot)}^{\text{word}}$ by a pre-trained model. 
%which is based on the cosine similarity. 

Then, an optimal transport plan matrix $\mathbf{T}^{*}(\mathbf{D},\mathbf{s}^*)=(t^{*}_{\alpha\alpha'}(\mathbf{D},\mathbf{s}^*))$ in pursuit of minimizing the transportation cost can be obtained by solving the following optimization problem: 
\begin{equation} \label{eqt:OT}
\begin{aligned}
\mathbf{T}^{*}(\mathbf{D},\mathbf{s}^*) = \underset{\mathbf{\mathbf{T}(\mathbf{D},\mathbf{s}^*)}}{\text{argmin}} \sum_{\alpha,\alpha'} t_{\alpha\alpha'}(\mathbf{D},\mathbf{s}^*)c_{\alpha\alpha}, \\
\text{s.t.} \; \sum_{\alpha'}t_{\alpha\alpha'}(\mathbf{D},\mathbf{s}^*) = \text{TF}_{\mathbf{D}}(\alpha),  \\
\sum_{\alpha=1}t_{\alpha\alpha'}(\mathbf{D},\mathbf{s}^*)t_{ij}^{Doc} =\text{TF}_{\mathbf{s}^{*}}(\alpha'), \\
t_{\alpha\alpha'}(\mathbf{D},\mathbf{s}^*)\geq 0, 
\forall \alpha, \alpha'.
\end{aligned}
\end{equation} 
To this end, the Wasserstein distance can be defined as:
\begin{equation}
\begin{aligned}
 L_{\text{document}} = \sum_{\alpha,\alpha'} t^{*}_{\alpha\alpha'}(\mathbf{D},\mathbf{s}^*)c_{\alpha\alpha'},
\end{aligned}
\end{equation} 
which is associated with the optimal transport plan. By minimizing $L_{\text{document}}$, a high-quality summary sentence is expected to be obtained. 

%The loss of textual coverage $\{L_{textual\; coverage}\}$ is computed by the Wasserstein distance   measuring the distance between the two semantic distributions $\text{TF}_{\mathbf{D}}$ and $\text{TF}_{\mathbf{S}}$ with the optimal transport plan:
%\begin{equation}
%\begin{aligned}
% L_{textual\; coverage} = \sum_{i,j} t^{*}_{text_{ij}}c_{ij} \;.
%\end{aligned}
%\end{equation} 

\subsubsection{Video Coverage}

In parallel, a good cover frame is supposed to be close to the original video regarding their perceptual similarity. We measure the loss of visual coverage by computing the Wasserstein distance $L_{\text{video}}$ between the corresponding colour signatures of the mean of video frames in $\mathbf{V}$ and the cover frame $\mathbf{f^*}$. It can be viewed as the minimal cost required to transport the semantics from $\mathbf{f^*}$ to $\mathbf{V}$. 

By denoting $\bar{\mathbf{f}}$ as the mean of the video frames in $\mathbf{V}$, %i.e. $\mathbf{f}_{v} = \frac{\mathbf{f}_{1} + ... + \mathbf{f}_{m}}{m}$. 
we define $\bar{\mathbf{r}}$ and $\mathbf{r}^{*}$ as the colour signatures of $\bar{\mathbf{f}}$ and $\mathbf{f^*}$, respectively. In detail, we have:
\begin{equation} \label{eqt:coloursig}
\begin{aligned}
\bar{\mathbf{r}} &= \{(\bar{\mu}_1, \bar{\tau}_1),... , (\bar{\mu}_{\bar{n}}, \bar{\tau}_{\bar{n}})\}\;, \\
\mathbf{r}^* &= \{(\mu^*_1, \tau^*_1), ..., (\mu^*_{m^*}, \tau^*_{m^*})\}\;,
\end{aligned}
\end{equation} 
where $\bar{\mu}_i$ and $\mu^*_j$ are the points in the colour space, and $\bar{\tau}_i$ and $\tau^*_j$ are the corresponding weights of the points.

An optimal transport plan matrix $\mathbf{T}^{*}(\mathbf{V},\mathbf{f}^*)=(t^{*}_{\beta\beta'}(\mathbf{V},\mathbf{f}^*))\in\mathbf{R}^{\bar{m}\times m^*}$ in pursuit of minimizing the transportation cost between $\bar{\mathbf{r}}$ and $\mathbf{r}^{*}$ can be obtained by solving the following optimization problem: 
\begin{equation} \label{eqt:OTvis}
\begin{aligned}
\mathbf{T}^{*}(\mathbf{V},\mathbf{f}^*) = \underset{\mathbf{T}(\mathbf{V},\mathbf{f}^*)}{\text{argmin}} \sum_{\beta, \beta'} t_{\beta\beta'}(\mathbf{V},\mathbf{f}^*)\left \| \bar{\mu}_\beta - \mu_{\beta'}^* \right \|\;, \\
\text{s.t.} \; \sum_{\beta'}t_{\beta\beta'}(\mathbf{V},\mathbf{f}^*) = \bar{\tau}_\beta, 
\sum_{\beta}t_{\beta\beta'}(\mathbf{V},\mathbf{f}^*) = \tau^*_{\beta'}\;, \\
t_{\beta\beta'}(\mathbf{V},\mathbf{f}^*)\geq 0, \forall \beta,  \beta'\;,
\end{aligned}
\end{equation} 
where $\mathbf{T}(\mathbf{V},\mathbf{f}^*)$ is a transport plan. Then, a Wasserstein distance measuring the distance between the two colour signatures can be derived as:
\begin{equation}
\begin{aligned}
 L_{\text{video}} = t^{*}_{\beta\beta'}(\mathbf{V},\mathbf{f}^*)\left \| \bar{\mu}_\beta - \mu_{\beta'}^* \right \|,%c_{ij},
\end{aligned}
\end{equation} 
which is associated with the optimal transport plan. By minimizing $L_{\text{video}}$, a high-quality summary frame is expected to be the cover frame. 

%The loss of visual coverage $\{L_{visual\; coverage}\}$ is computed by the Wasserstein distance  measuring the distance between the two colour signatures  $\mathbf{cs}_{\mathbf{f}_{v}}$ and $\mathbf{cs}_{\mathbf{f^*}}$ with the optimal transport plan:
%\begin{equation}
%\begin{aligned}
% L_{visual\; coverage} = \sum_{i,j} t^{*}_{vis_{ij}} \;.
%\end{aligned}
%\end{equation} 

\subsection{Textual Fluency and Cross-modal Consistency}

Inspired by \citet{PLaban2020SummaryLoop}, we adopt a pre-trained language model $P_{LM}$ to measure 
the fluency of the textual summary 
\(L_{\text{Fluency}}\). The loss can be defined as: 
%Given a trained language model \(LM\), \(\text{L}_{\text{flu}}\) of a summary $\mathbf{S}$ is defined as: 
\begin{equation}
L_{\text{Fluency}}= P_{LM}(\mathbf{s}^*),
\end{equation} 
where \(P_{LM}\) computes the probability of $\mathbf{s}^*$ being a sentence. 

%\subsection{Cross-modal Consistency}

%Intuitively, a good multimodal summary is supposed to have close 
The semantic consistency should exist between the cover frame and the one-sentence summary. %indicates the semantic consistency across two modalities.  
To formulate this, we measure the cross-modal similarity between the two embeddings of the cover frame $\mathbf{f}^*$ and the one-sentence summary $\mathbf{s}^{*}$. The loss can be defined based on a cosine similarity:
\begin{equation}
\begin{aligned}
 L_{\text{cross-modal}} = 1 - cos(\mathbf{f}^*, \mathbf{\mathbf{s}^*}).
\end{aligned}
\end{equation}

In summary, four losses have been obtained to measure the summarisation quality: $ L_{\text{document}}$, $L_{\text{video}}$, $L_{\text{fluency}}$ and $L_{\text{cross-modal}}$. To this end, a loss function to optimize the proposed architecture can be formulated as follows:
\begin{equation}
\begin{aligned}
L = \lambda_{\text{d}}L_{\text{document}} + \lambda_{\text{v}}L_{\text{video}} 
+  \lambda_{\text{f}}L_{\text{fluency}} 
+\lambda_{\text{c}}L_{\text{cross-modal}},
\end{aligned}
\end{equation}
where $\lambda_{\text{d}}$, $\lambda_{\text{v}}$, $\lambda_{\text{f}}$ and $\lambda_{\text{c}}$ are the hyper-parameters controlling the weights of each loss term.

\section{Experimental Results and Discussions}

\subsection{Dataset}
To the best of our knowledge, there is no existing large-scale dataset for XMSMO. Hence, we collected the first large-scale dataset of such kind, XMSMO-News, from the British Broadcasting Corporation (BBC) News Youtube channel \footnote{https://www.youtube.com/c/BBCNews}. We used the Pytube library to collect 4,891 quartets of video, document, cover frame, and one-sentence summary from the year 2013 to 2021. We used the video description as the document and video title as the one-sentence summary, as these visual and textual summaries were professionally created by the BBC. \footnote{We removed the trailing promotional text from the video title and video description.} We then split the quartets randomly into the train, validation, and test sets at a ratio 90:5:5.

Table \ref{tab:dataset} shows the statistics and the comparison of XMSMO-News with other benchmarks on multimodal summarisation with multimodal output. %XMSMO-News shares some similarity with the dataset proposed in \cite{JZhu2018msmo,MLi2020vmsmo} in terms of the input and output modalities. However, 
The major differences are regarding the input and output lengths: XMSMO-News has an average duration of 345.5 seconds, whereas \cite{MLi2020vmsmo} has 60 seconds only. %In addition, our dataset is in English, whereas \cite{MLi2020vmsmo} is in Chinese. 

\begin{table}[htbp]
\tiny
  \centering
 \resizebox{.475\textwidth}{!}
 {
 \begin{threeparttable}
    \begin{tabular}{p{2.5cm}|p{2.5cm}<{\centering}p{2.8cm}<{\centering}p{2.9cm}<{\centering}}
    %\hline
    \textbf{Dataset} & \textbf{XMSMO-News} & \textbf{VMSMO}  & \textbf{MSMO}   \\
    \hline\hline
    \text{\#Train/Val/Test} & 4382/252/257 & 180000/2460/2460 &  293965/10355/10262  \\
    %\hline
    \text{Language} & English & Chinese & English   \\
    %\hline
    \text{Visual Input} & Video & Video & Multi-images   \\
    %\hline
    \text{Textual Input} &  Document &  Document & Document   \\
    %\hline
    \text{Visual Output} & Cover frame  & Cover frame & One image \\
    %\hline
    \text{Textual Output} & One-sentence & Arbitrary length & Multi-sentence   \\
    %\hline
    
    \text{Frames/Video} & 8827.4 & 1500.0  & 6.6 \\
    %\hline
    \text{Video Duration(s)} & 345.5  & 60.0 & -   \\
    %\hline
    \text{Tokens/Document} & 101.7  & 96.8 & 723.0\\
    %\hline
    \text{Tokens/Summary} & 12.4  & 11.2 & 70.0 \\
    %\hline
    \text{Annotation} & Full & Partial\tnote{1}   & Partial\tnote{2} \\
    %\hline
    \end{tabular}%
      \begin{tablenotes}
    \item[1,2] 1) Not all ground-truth data is available; 2) No visual ground-truth on training and validation splits.
  \end{tablenotes}
    \end{threeparttable}
    }
    \caption{Comparison of XMSMO-News with existing MSMO benchmark datasets. }
  \label{tab:dataset}%
\end{table}%

\subsection{Implementation Details}

We used the PyTorch library for the implementation of our method. We set the hidden size of GPO and GRU to 512. 
For the pre-trained CLIP model and the pre-trained token embedding model BERT (base version) used for computing the loss of textual coverage, we obtained them from HuggingFace \footnote{\label{huggingface}https://huggingface.co}. To detect the scenes of a video, we utilised the PySceneDetect library \footnote{http://scenedetect.com/en/latest/}. 
%To compute the Wasserstein distances, we utilised the POT library \footnote{https://pythonot.github.io} and the OpenCV library, respectively.  
For video preprocessing, we extracted one of every 360 frames to obtain 120 frames as candidate frames. All frames were resized to 640x360. 
We trained HOT-Net using AdamW \cite{ILoshchilov2018AdamW} with a learning rate of 0.01 and a batch size of 3 for about 72 hours. All experiments were run on a GeForce GTX 1080Ti GPU card. %For evaluation, We obtained our ROUGE scores by using the pyrouge package \footnote{https://pypi.org/project/pyrouge/}.

\subsection{Baselines}
To evaluate our proposed method HOT-Net, we compared it with the following state-of-the-art baseline methods, including PEGASUS \cite{JZhang2020pegasus} - the state-of-the-art method of text summarisation, CA-SUM  \cite{EApostolidis2022CASUM}  - the state-of-the-art method of video summarisation, zero-shot CLIP \cite{ARadford2021CLIP} - the state-of-the-art multi-modal embedding model with a linear classification layer to perform multimodal summarisation. The baseline models PEGASUS and CLIP were obtained from HuggingFace %\footnotemark[\ref{huggingface}]
; CA-SUM was obtained from the author's Github \footnote{https://github.com/e-apostolidis/CA-SUM}; VMSMO was obtained from the author's Github \footnote{https://github.com/iriscxy/VMSMO} with modifications on the latest libraries' update and bug fixing.

\begin{table*}[h]
\footnotesize
\centering
\resizebox{\textwidth}{!}{
\begin{tabular}{l|ccc|cc|c}
%\hline
\textbf{Method} &  \multicolumn{3}{c|}{\textbf{Textual Evaluation}}
& \multicolumn{2}{c|}{\textbf{Visual Evaluation}} & \textbf{Overall Evaluation}  \\
 {} & ROUGE-1 & ROUGE-2 & ROUGE-L  & Frame Accuracy & IoU  & {}\\ \hline\hline
% XSum \cite{SNarayan2018XSum} & 5.44 & \textbf{0.16} & \ul{5.01} & - & -  \\
PEGASUS \cite{JZhang2020pegasus} & 4.36 & \textbf{0.12} & \ul{4.00} & - & -  & - \\
CA-SUM \cite{EApostolidis2022CASUM} & - & - & - & \ul{0.57} & \ul{0.69} & - \\
VMSMO \cite{MLi2020vmsmo} & \textit{Divergence} & \textit{Divergence} & \textit{Divergence} %\tablefootnote{The model generates "UNK" tokens as the textual summary} 
& 0.57 & 0.69   & 0.49
\\
CLIP \cite{ARadford2021CLIP} & 4.14 & \ul{0.08} & 3.80 & 0.54 & 0.63  & 0.89 \\\hline\hline
HOT-Net (Ours) visual only & - & - & - & \textbf{0.60} & 0.68 & - \\
HOT-Net (Ours) textual only & 3.85 & 0.05 & 3.60 & - & - & - \\
%These two are used to compare the effectiveness to multimodal learning
HOT-Net (Ours) w/o multimodal fusion & 3.99 & 0.05 & 3.73 & 0.56 & \textbf{0.70} & 0.93  \\
HOT-Net (Ours) w/o local-level multimodal fusion   &  4.45 & 0.06 & 4.16 & 0.59 & \textbf{0.70} & 0.98 \\
HOT-Net (Ours) w/o global-level multimodal fusion \hspace{5cm}   & 3.65 & 0.06 & 3.45 & 0.58 & 0.68 & 0.88 \\ 
HOT-Net (Ours) w/o fluency loss   & 4.58 & 0.06 & 4.28 & 0.57 & 0.68 & 0.98 \\
HOT-Net (Ours) w/o cross-modal loss  & 4.58 & 0.06 & 4.28 & 0.57 & 0.68 & 0.98  \\
\hline\hline
HOT-Net (Ours)   & \textbf{4.64} & 0.07 & \textbf{4.33} & \ul{0.57} & 0.68 & \textbf{0.99} \\
%\hline

\end{tabular}
}
\caption{Comparisons between our HOT-N and the state-of-the-art summarisation methods on XMSMO-News.}
%\footnotesize{The VMSMO model generates "UNK" tokens as the textual summary.}\\
\label{tab:evaluation}
\end{table*}

\subsection{Quantitative Analysis} 
For the quantitative evaluation of a textual summary, the commonly used ROUGE metric \cite{CLin2004Rouge} for text summarisation is adopted. For the visual summary, the commonly used Intersection over Union (IoU) \cite{ASharghi2017Query} and frame accuracy \cite{SMessaoud2021DeepQAMVS} metrics for video summarisation are adopted. 

The ROUGE metric evaluates the content consistency between a generated summary and a reference summary. In detail, the ROUGE-n F-scores calculates the number of overlapping n-grams between a generated summary and a reference summary. The ROUGE-L F-score considers the longest common subsequence between a generated summary and a reference summary. 
IoU metric evaluates the high-level semantic information consistency by counting the number of overlap concepts between the ground-truth cover frame and the generated one. Frame accuracy metric is to compare lower-level visual features, the ground-truth cover frame and generated cover frame are considered to be  matching when pixel-level Euclidean distance is smaller than a predefined threshold. %\footnote{We followed \cite{SMessaoud2021DeepQAMVS} to set the predefined threshold to 0.6.} 
To evaluate the overall performance on both modalities, we compute the overall evaluation as $0.5 \times \frac{\text{IoU}}{\text{Best IoU}} + 0.5 \times \frac{\text{ROUGE-L}}{\text{Best ROUGE-L}} $, where the best IoU and the best ROUGE-L are the best scores among all the evaluated methods.

The experimental results of HOT-Net on XMSMO-News are shown in Table \ref{tab:evaluation} 
including ROUGE-1, ROUGE-2. and ROUGE-L F-scores, and IoU. Our method outperforms the baseline models in terms of ROUGE-1 and ROUGE-L, which demonstrate the quality of the generated extreme textual summary, and achieves promising results in terms of frame accuracy and IoU, which demonstrate the quality of the generated extreme visual summary. HOT-Net underperforms in terms of ROUGE-2, which may be due to the trade-off between informativeness and fluency. PEGASUS was trained on massive text corpora which may help improve the fluency of natural language generation. This trade-off is further discussed in the Qualitative Analysis section.

\begin{figure}[h!]
\centering
\includegraphics[width=.47\textwidth]{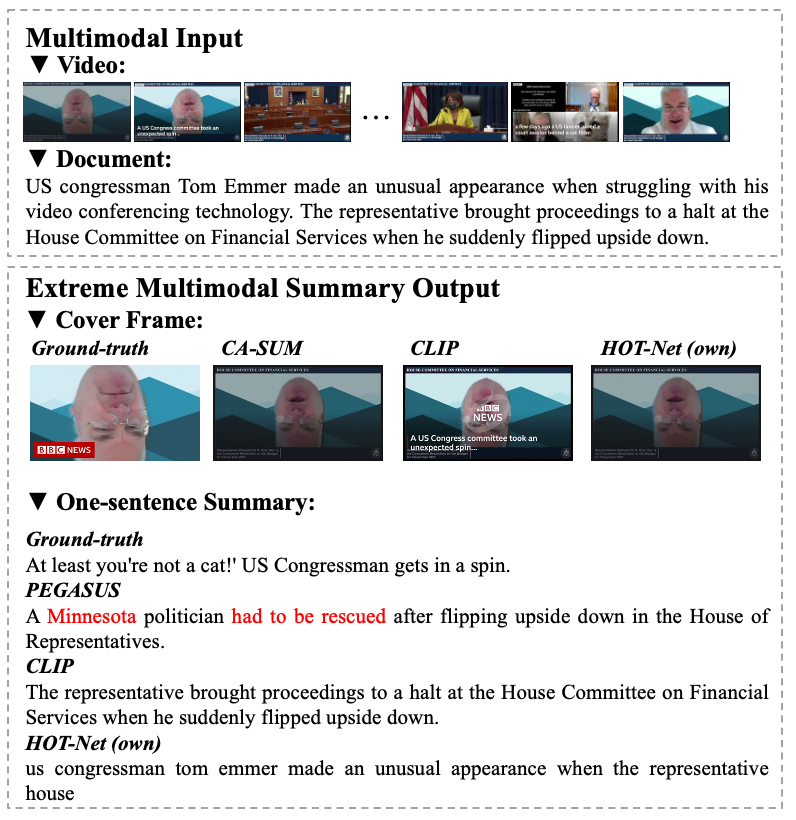}
\caption{Example summary generated by baseline methods and HOT-Net on XMSMO-News. It is about a US congressman made an unusual appearance and flipped upside down. %HOT-Net produces factually correct and reasonably fluent extreme textual summary that captures the essence of the document. In comparison, as highlighted in red colour, PEGASUS produces fluent yet unfaithful summary. Most of the methods agree on the choice of cover frame which may due to its visual representativeness.
}
\label{fig:sample}
\end{figure}

\subsubsection{Ablation Study}

To study the effect of the proposed mechanisms, we compare a number of different settings of HOT-Net and the results can be found in Table 2. We first observe that multimodal learning improves the modelling by comparing to the visual or textual only method. Our fusion strategy is also important to obtain high-quality textual summaries. The hierarchical mechanism does not have much impact on the results of the visual summary, which may be due to that the overall model architecture has achieved its best possible potential in terms of producing a visual summary. Additionally, the fluency loss and cross-modal loss improve the textual summary as well.

\subsection{Qualitative Analysis}

Figure \ref{fig:sample} compares the summaries produced by HOT-Net and the baseline methods, and the reference summary of a sample in the XMSMO-News dataset.  
The example demonstrates that our proposed HOT-Net method produces factually correct and reasonably fluent extreme textual summary that captures the essence of the document even without supervision. In comparison, as highlighted in red colour, PEGASUS produces a fluent but unfaithful summary with information that does not occur in the original document. 
Most of the methods agree on the choice of the cover frame, whilst ours and CA-SUM are closer to the ground-truth. %For the second example, since the aeroplane appears repeatedly and occupies comparatively large area on the frames, there is room for improvement to learn and identity the information which human considers to be \textit{important}, such as a frame containing the face of the key human figure.  

\section{Conclusion}

In this paper, we have introduced a new task - eXtreme Multimodal Summarisation with Multimodal Output (XMSMO), which aims to summarise a video-document pair into an extreme multimodal summary, consisting of one cover frame as the visual summary and one sentence as the textual summary. We present a novel {\it unsupervised} deep learning architecture, which consists of three components:  hierarchical multimodal encoders,  hierarchical multimodal fusion decoders, and  optimal transport solvers.  %To achieve unsupervised learning, besides the optimal transport based semantic coverage guidance,  textual fluency and cross-modal similarity are explored as well. 
In addition, we construct a new large-scale dataset XMSMO-News to facilitate research in this new direction. Experimental results demonstrate the effectiveness of our method. 
In the future, we will explore the metric space to measure the optimal transport plan in a more efficient and effective manner. Moreover, we will explore improved ways to learn and identity the information that humans would consider to be \textit{important}, such as a frame containing the face of a key character.

\label{cha:conclusion}

%\clearpage

\bibliography{custom}

\end{document}